\documentclass[letterpaper, 10 pt, conference]{ieeeconf}  

\IEEEoverridecommandlockouts                              

\usepackage{graphics} 
\usepackage{epsfig} 
\usepackage{mathptmx} 
\usepackage{times} 
\usepackage{amsmath} 
\usepackage{amssymb}  
\usepackage{comment}
\usepackage[T1]{fontenc}
\usepackage{multirow}
\usepackage{textcomp}
\usepackage{gensymb}
\usepackage{balance}
\usepackage{soul}
\usepackage{color}

\usepackage{adjustbox}
\usepackage[utf8]{inputenc}
\usepackage{csquotes}
\usepackage[english]{babel}

\usepackage[backend=biber,style=numeric,sorting=none, maxnames=99, giveninits=true]{biblatex} 
\title{\LARGE \bf
Localization and Mapping of Sparse Geologic Features with Unpiloted Aircraft Systems
}

\author{Zhiang Chen, Sarah Bearman, J Ram\'{o}n Arrowsmith, Jnaneshwar Das
\thanks{Authors are affiliated with the School of Earth and Space Exploration, Arizona State University, 781 Terrace Mall, Tempe, AZ 85287, USA
        }%
}

\begin{document}

\maketitle
\thispagestyle{empty}
\pagestyle{empty}
\maxdeadcycles=200

\begin{abstract}
Robotic mapping is attractive in many scientific applications that involve environmental surveys. This paper presents a system for localization and mapping of sparsely distributed surface features such as precariously balanced rocks (PBRs), whose geometric fragility parameters can provide valuable information on earthquake processes and landscape development. With this geomorphologic problem as the test domain, we carry out a lawn-mower search pattern from a high elevation using an Unpiloted Aerial Vehicle (UAV) equipped with a flight controller, GPS module, stereo camera, and onboard computer. Once a potential PBR target is detected by a deep neural network in real time, we track its bounding box in the image coordinates by applying a Kalman filter that fuses the deep learning detection with Kanade–Lucas–Tomasi (KLT) tracking. The target is localized in world coordinates using depth filtering where a set of 3D points are filtered by object bounding boxes from different camera perspectives. The 3D points also provide a strong prior on target shape, which is used for UAV path planning to closely map the target using RGBD SLAM. After target mapping, the UAS resumes the lawn-mower search pattern to locate and map the next target. 
\end{abstract}

\section{introduction}
Although topographic mapping using Unpiloted Aerial Systems (UAS) and Structure from Motion (SfM) has recently proliferated in large-scale terrain mapping, UAS-SfM methods are not ideal when the mapping is designed to map and identify sparsely distributed targets of interest. This is because UAS-SfM methods based on bundle adjustment require intensive offboard computation to produce dense terrain maps that include both targets of interest and other irrelevant features. As the sparsity of targets increases, significant amounts of computation and memory storage are spent on reconstructing irrelevant features. More importantly, UAS-SfM methods like offboard mapping decouple robot navigation from mapping. This inhibits adaptive or active mapping~\cite{bircher2016receding, delmerico2017active} that can optimize flight paths to locate and map specific targets. 

Target localization is a process of estimating a target’s location in 3D space. Previously, target location was simplified by determining the target’s geometric center, which is useful when the target’s dimensions are negligible. However, in scenarios where the target’s dimensions matter, a representation of the target’s geometric shape information is desired. In this work, a target’s location is represented by a set of 3D points, which can indicate its geometric center, orientation, and dimension. The additional information facilitates UAV path planning for target mapping.

\begin{figure}
\centering
\vspace{6pt}
\includegraphics[width=0.48\textwidth]{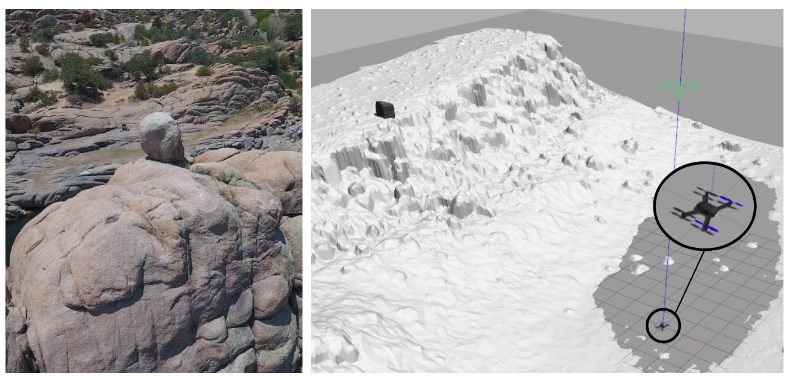}
\caption{Precariously balanced rocks: (Left) PBR in Granite Dell, Arizona. (Right) PBR mapping simulation environment in Gazebo. In simulation, PBR (with dark texture at the top-left in right figure) is on the foot wall of a fault scarp. }
\label{fig:pbrs}
\end{figure}

This work is motivated by the challenges of finding and mapping precariously balanced rocks (PBRs) (Fig.~\ref{fig:pbrs}) and other fragile geologic features (e.g. rock pillars). PBRs are easily toppled by strong earthquake shaking, so their existence can provide valuable information about earthquake history in a given region~\cite{brune1996precariously, anooshehpoor2004methodology, haddad2012estimating}. 

However, model assessment in a broad statistical sense is challenging because PRBs are sparsely distributed and difficult to identify. Our ultimate goal is to assess the precarious features broadly, increasing the number of analyses and accounting for their geomorphic setting~\cite{anooshehpoor2004methodology, haddad2012applications}. Large datasets of spatially explicit PBR fragility with detailed geomorphic and geologic context provide a valuable assessment of sensitivity to ground motions. Instead of having a few fragilities, it should be possible to have a fragility histogram for the fragile geologic features at a site.

\begin{figure*} [tp]
\centering
\vspace{6pt}
\includegraphics[width=0.9\textwidth]{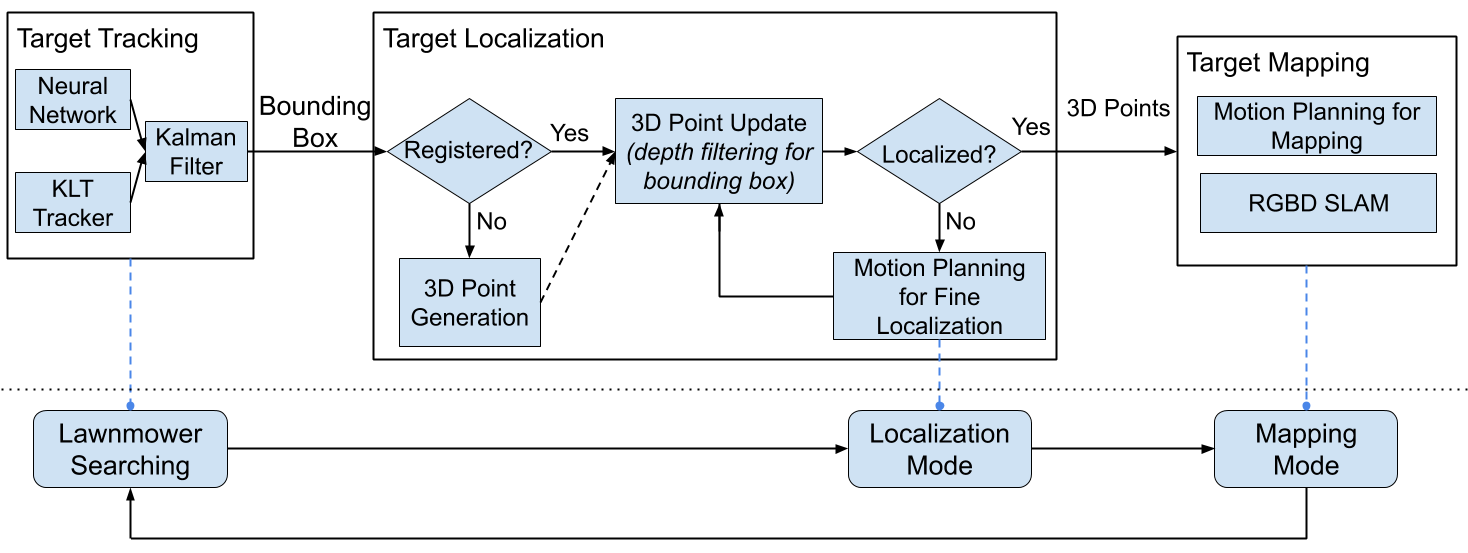}
\caption{System workflow: The system includes target tracking, target localization, and target mapping. The motion mode switches from lawn-mower searching during target tracking to localization mode once fine localization is requested, and then switches to mapping mode after the target is finely localized.}
\label{fig:workflow}
\end{figure*}

The objective of this paper is to present an efficient, real-time approach to localize and map PBRs by taking advantage of deep neural network object detection, sample-based state estimation, and RGBD SLAM. The system workflow is presented in Fig.~\ref{fig:workflow}. We start by tracking bounding box coordinates in image frames using a Kalman filter that fuses deep neural network detection and KLT tracking. Then, the target is localized in world coordinates by sampling a set of 3D points that are generated by back-projecting a polyhedral cone from the vertices of the enlarged tracked bounding box at different camera perspectives. Finally, we deploy RGBD SLAM for accurate target mapping. For the convenience of tracking, the math notations are consistent in the entire paper.

\section{target tracking}

We employ a deep neural network to detect targets, and then track their bounding boxes in image coordinates using a Kalman filter. A KLT tracker is used to predict the new state of each bounding box, and the state variable is updated using object detection from a well-trained deep neural network. We use an IoU threshold to register new bounding boxes, and we deregister the bounding boxes when they move out of the current image frame. False detections are deregistered by examining the differential entropy of bounding box state distributions.

The Kalman filter algorithm is presented in Algorithm 1. The state variable of the Kalman filter is the bounding box coordinates  $\mathbf{x}=(u_{min}, v_{min}, 1, u_{max}, v_{max}, 1)$ in homogeneous coordinate system, or $\mathbf{u}=(u_{min}, v_{min}, u_{max}, v_{max})$ in Euclidean coordinate system. Homogeneous coordinate system and Euclidean coordinate system can be converted by matrix operations:
$$\mathbf{x}_{t-1} = \mathbf{T}\mathbf{u}_{t-1} + \mathbf{a}$$
$$\mathbf{u}_t = \mathbf{M}\mathbf{x}_t$$
where \\
$\mathbf{T} = 
\begin{bmatrix}
1 & 0 & 0 & 0  \\
0 & 1 & 0 & 0  \\
0 & 0 & 0 & 0  \\
0 & 0 & 1 & 0  \\
0 & 0 & 0 & 1  \\
0 & 0 & 0 & 0  \\
\end{bmatrix}
$,
$\mathbf{a} = 
\begin{bmatrix}
0   \\
0   \\
1   \\
0   \\
0   \\
1   \\
\end{bmatrix}
$,
$\mathbf{M} = 
\begin{bmatrix}
1 & 0 & 0 & 0 & 0 & 0 \\
0 & 1 & 0 & 0 & 0 & 0 \\
0 & 0 & 0 & 1 & 0 & 0 \\
0 & 0 & 0 & 0 & 1 & 0 \\
\end{bmatrix}
$ 
Lines 1-2 and Lines 5-6 in Algorithm correspond the conversions between homogeneous and Euclidean coordinates.

The predict phase uses a KLT tracker to estimate a bounding box state. We consider a linear prediction model for bounding boxes,
\begin{equation}\label{predict_model}
\mathbf{x}_{t}=\mathbf{F}_{t-1}\mathbf{x}_{t-1}+\mathbf{e}_{t-1}
\end{equation}
where $\mathbf{F}_{6\times6}$ is state transition model, and $\mathbf{e}\in\mathbb{R}^6$ is a prediction error vector:
$$\mathbf{F} = 
\begin{bmatrix}
\mathbf{S} & \mathbf{0} \\
\mathbf{0} & \mathbf{S}
\end{bmatrix}, \mathbf{e} = (e, e, 0, e, e, 0)$$
$\mathbf{S}_{3\times3}$ is a similarity transform produced from KLT tracker that applies Harris features in each bounding box. We use homogeneous coordinate system in the predict phase to keep consistent with $\mathbf{F}$, which is in 2D homogeneous coordinates. 

{\small 
\begin{table} 
\resizebox{\columnwidth}{!}{%
\begin{tabular}{l}
\hline
\textbf{Algorithm 1} Kalman Filter for Bounding Box Tracking \\ 
\hline
\begin{tabular}[c]{@{}l@{}}
\textbf{input}: $\mathbf{u}_{t-1}$, $\Sigma_{t-1}$, $\mathbf{F}_{t-1}$, $\mathbf{z}_t$\\ 
\textbf{output}: $\mathbf{u}_t$ \\
\textbf{parameter}: $\mathbf{R}$, $\mathbf{Q}$
\end{tabular}\\ 
\hline
Predict: \\
1. \hspace{5mm} $\mathbf{x}_{t-1} = \mathbf{T}\mathbf{u}_{t-1} + \mathbf{a}$  \\
2. \hspace{5mm} $\Omega_{t-1} =  \mathbf{T}\Sigma_{t-1} + \mathbf{a} $\\
3. \hspace{5mm} $\mathbf{x}_{t} = \mathbf{F}_{t-1}\mathbf{x}_{t-1}$ \\
4. \hspace{5mm} $\Omega_t = \mathbf{F}_{t-1} \Omega_{t-1} \mathbf{F}_{t-1}^T + \mathbf{R}$ \\
5. \hspace{5mm} $\mathbf{u}_t = \mathbf{M}\mathbf{x}_t$  \\
6. \hspace{5mm} $\Sigma_t =  \mathbf{M}\Omega_t$\\
Update: \\
7. \hspace{5mm} $\mathbf{K} = \Sigma_t (\Sigma_t + \mathbf{Q})^{-1}$ \\
8. \hspace{5mm} $\mathbf{u}_t = \mathbf{u}_t + \mathbf{K} (\mathbf{z}_t - \mathbf{u}_t)$ \\
9. \hspace{5mm} $\Sigma_t = (\mathbf{I}_{4\times4} - \mathbf{K})\Sigma_t$\\
\hline
\end{tabular}}
\end{table}
}

Object detection from the deep neural network is deployed to update the state variable,
\begin{equation}\label{update_model}
\mathbf{z}_{t} = \mathbf{H}\mathbf{u}_{t} + \mathbf{w}_{t}
\end{equation}
where $\mathbf{H}_{4\times4}$ is an identity matrix, $\mathbf{w}\in\mathbb{R}^4$ is an error vector for the update model, and $\mathbf{z} = (u_{min}, v_{min}, u_{max}, v_{max}) \in\mathbb{R}^4$ is bounding box coordinates from the neural network. We obtain these coordinates from 
\begin{equation}
 \mathbf{z}_{t}=f(I_{t})  
\end{equation}
where $I$ is an input image and $f$ is a well-trained object detection neural network. We assume both  predict error $\mathbf{e}$ and update error $\mathbf{w}$ have multi-variable Gaussian distribution:
$$\mathbf{e}\sim\mathcal{N}(\mathbf{0}, \mathbf{R}),
\mathbf{w}\sim\mathcal{N}(\mathbf{0}, \mathbf{Q})$$
where $\mathbf{R}_{6\times6}$, $\mathbf{Q}_{4\times4}$ are the covariance matrices: \\
$$\mathbf{R} = 
\begin{bmatrix}
e^2 & 0 & 0 & 0 & 0 & 0 \\
0 & e^2 & 0 & 0 & 0 & 0 \\
0 & 0 & 0 & 0 & 0 & 0 \\
0 & 0 & 0 & e^2 & 0 & 0 \\
0 & 0 & 0 & 0 & e^2 & 0 \\
0 & 0 & 0 & 0 & 0 & 0
\end{bmatrix},
\mathbf{Q} = w^2 \mathbf{I}_{4\times4}
$$
and $\mathbf{I}_{4\times4}$ is an identity matrix. We use Euclidean coordinate system in the update phase because an invertible covariance matrix is needed to update Kalman gain $\mathbf{K}$, while the covariance matrix in homogeneous coordinate system $\Omega$ is degenerate. 


A new bounding box detected from the neural network will be registered if its IoU is smaller than a threshold with any previously registered bounding boxes. Otherwise, the new bounding box coordinates will be used to update the registered bounding boxes that satisfy the IoU threshold. A bounding box will be deregistered when it is moving out of the current image frame. 

\begin{figure}
\centering
\vspace{6pt}
\includegraphics[width=0.5\textwidth]{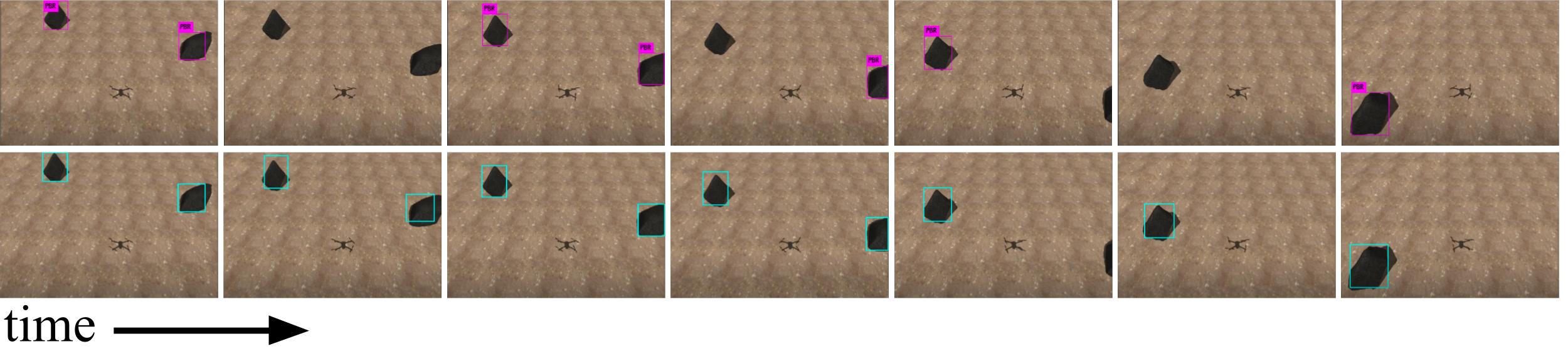}
\caption{Kalman filtering for bounding box tracking: The UAV is moving north. The top-row images are neural network detection. The bottom-row images are the results from the Kalman-filtering tracking.}
\label{fig:tracking}
\end{figure}

False detection from the deep neural network can be problematic, so we monitor the differential entropy of each bounding-box-state distribution: 
\begin{equation}
h=\frac{k}{2} + \frac{k}{2}\ln(2\pi) + \frac{1}{2}\ln(|\Sigma|)  
\end{equation}
where $k=4$ is the dimensionality of the state vector space. When the differential entropy $h$ becomes greater than a threshold, it implies there has been no update for several predictions, and the bounding box will be deregistered. An example of tracking results is shown in Fig.~\ref{fig:tracking}.

\section{target localization}
\begin{figure}
\centering
\vspace{6pt}
\includegraphics[width=0.50\textwidth]{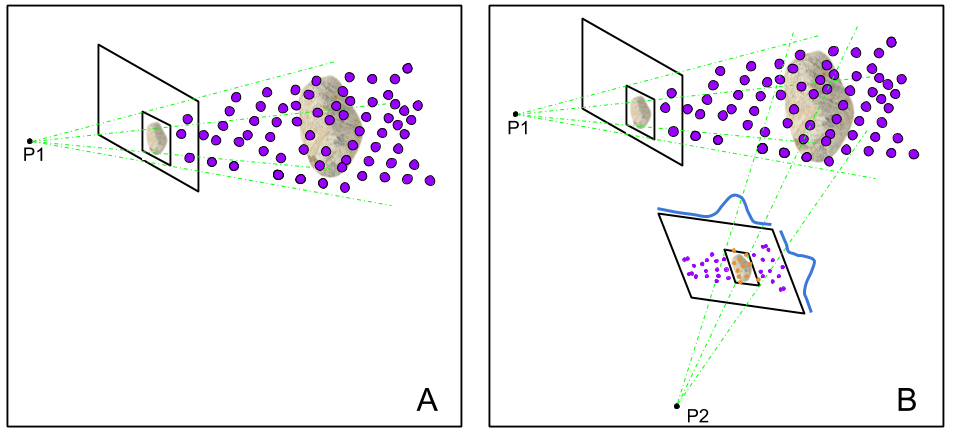}
\caption{3D point generation and update: (A) 3D points are generated within a polyhedral cone that is back-projected from an enlarged bounding box. (B) 3D points are projected to a different image frame and resampled according to their projections' positions with respect to the new bounding box.}
\label{fig:depth_filter}
\end{figure}

Once a target is tracked in image frames, we first randomly generate 3D points within a four-face polyhedral cone that is back-projected from an enlarged bounding box (Fig.~\ref{fig:depth_filter}(A)). The 3D points will be projected to different image frames and resampled according to their positions with respect to the tracked bounding box (Fig.~\ref{fig:depth_filter}(B)). The rationale is similar to the depth filter in \cite{Forster2014ICRA}, but we adapt it to bounding boxes and use a set of 3D points to represent a target location. Unlike other work that simplifies target locations as geometric centers, 3D points offer added information such as orientation and dimension, which can facilitate UAV motion planning for target mapping.

\subsection{Camera Model}
We derive the back-projection of image points to rays on special Euclidean group based on the models in Hartley and Zisserman's book \cite{hartley2003multiple}. A general projective camera maps a point in space $\mathbf{L}=(x, y, z, 1)$ to an image point $\mathbf{l}=(u, v, 1)$ according to the mapping $\mathbf{l} = \mathbf{P}\mathbf{L}$. $\mathbf{P}$ is the projection matrix, and $\mathbf{P}=[K|\mathbf{0}]T_c^w$, where $K_{3\times3}$ is the camera intrinsic matrix, and $T_c^w\in SE(3)$ is the transformation matrix from world frame to camera frame. We can back-project a ray from an image point,
\begin{equation}
    \mathbf{r}_c = S(\mu K^{-1}\mathbf{l}) + \mathbf{b}
\end{equation}
\begin{equation}
    \mathbf{r}_w = T_w^c \mathbf{r}_c
\end{equation}
where $\mathbf{r}_c$ is a homogeneous 3D point representing a ray in camera frame, $\mathbf{r}_w$ is the corresponding ray in world frame, $\mu$ is a non-negative scalar, $S_{4\times3} = \begin{bmatrix}
I_{3\times3} \\
\mathbf{0}
\end{bmatrix}$, $\mathbf{b} = (0, 0, 0, 1)$, and $T_w^c\in SE(3)$ is the transformation matrix from camera frame to world frame. $S$ and $\mathbf{b}$ convert Euclidean coordinates to Homogeneous coordinates. A ray in world frame is represented by a 3D points set $\eta=\{\mathbf{r}_w| \mu \geq0 \}$.

\subsection{3D Point Generation}
We have $T_w^c = T_w^b T_b^c$, where $T_b^c$ is the transformation matrix from camera frame to robot base frame, and $T_w^b$ is the transformation matrix from the robot base frame to the world frame. We enlarge the bounding boxes from target tracking to account for errors in the drone state estimator. A four-face polyhedral cone, $\{\eta_{w1}, \eta_{w2}, \eta_{w3}, \eta_{w4}\}$, is generated by four 3D rays back-projected from four corners of the enlarged bounding boxed. To randomly initialize a 3D point in the cone, we adopt the following steps.
\begin{enumerate}
\item randomly generate a direction vector from a convex combination of four enlarged corner vectors:
\begin{equation}
    \mathbf{v} = \alpha_1 K^{-1} \mathbf{l}_1 + \alpha_2 K^{-1} \mathbf{l}_2 + \alpha_3 K^{-1} \mathbf{l}_3 + \alpha_4 K^{-1} \mathbf{l}_4
\end{equation}
s.t. \\
$\alpha_1, \alpha_2, \alpha_3, \alpha_4 \sim \mathcal{U}$ \\
$\alpha_1 + \alpha_2 + \alpha_3 + \alpha_4 = 1$ \\
where $\mathcal{U}$ is a standard uniform distribution, $\{\mathbf{l}_i\}$  are corners of an enlarged bounding box in image frame.
\item
randomly scale the direction vector: $\mathbf{\Bar{L}} = \delta\mathbf{v}$, where $\delta$ is a random depth. $\mathbf{\Bar{L}}$ is a Homogeneous 3D point in the cone in camera frame.
\item
convert the 3D point to world frame: $\mathbf{L}=T_w^c(S\mathbf{\Bar{L}} + \mathbf{b})$.  
\end{enumerate}
$\mathbf{L}$ is one randomly generated point within a cone in world frame. We generate 1000 3D points to localize a tracked target using the matrix form of 3D point generation in Algorithm 2.

{\small 
\begin{table} 
\vspace{6pt}
\resizebox{\columnwidth}{!}{%
\begin{tabular}{l}
\hline
\textbf{Algorithm 2} 3D Point generation \hspace{20mm} \\ 
\hline
\begin{tabular}[c]{@{}l@{}}
\textbf{input}: $\{\mathbf{l}_1, \mathbf{l}_2, \mathbf{l}_3, \mathbf{l}_4\}$, $m$, $T_w^c$ \\ 
\textbf{output}: $P_w$ \\
\textbf{parameter}: $K$, $d_m$
\end{tabular}\\ 
\hline
1. \hspace{5mm} $A = \mathbb{U}(4, m)$  \\
2. \hspace{5mm} $B = Diag(|A|_m)$  \\
3. \hspace{5mm} $\Bar{A} = AB^{-1}$  \\
4. \hspace{5mm} $C = [K^{-1}\mathbf{l}_1, K^{-1}\mathbf{l}_2, K^{-1}\mathbf{l}_3, K^{-1}\mathbf{l}_4]$  \\
5. \hspace{5mm} $\delta = d_m\mathbb{U}(1,m)$  \\
6. \hspace{5mm} $\Bar{\delta}=Diag(\delta)$  \\
7. \hspace{5mm} $\Tilde{P}_c = C\Bar{A}\Bar{\delta}$  \\
8. \hspace{5mm} $P_c = S\Tilde{P}_c + \mathbf{b}$ \\
9. \hspace{5mm} $P_w = T_w^c P_c$  \\
\hline
Comments: \\
1). $m$ is the number of 3D points to be generated, e.g. $m=1000$.\\ 
2). $\mathbb{U}(x, y)$ is a standard uniform sampler returning a \\$x\times y$ matrix. \\
3). $Diag(x)$ creates a diagonal matrix from a vector $x$. \\
4). $|\cdot|_m$ returns a vector of Manhattan distances of all columns. \\
5). $d_m$ is the maximum depth of 3D points along $z$ axis \\in camera frame. \\
\hline
\end{tabular}}
\end{table}
}

When a bounding box is updated in tracking, a new set of 3D points will be generated when there is no existing 3D point in the cone back-projected from the updated bounding box. The four-face polyhedral cone from a bounding box can be represented by four inward half-spaces. For convenience, we use camera frame for 3D point registration because all four hyperplanes pass through the origin. Then we have a 3D point registration condition:
\begin{equation}
\mathbf{N}\mathbf{\Tilde{L}} > \mathbf{0}
\end{equation}
where $\mathbf{\Tilde{L}} = (x, y, z)$ is the 3D point in camera frame, $\mathbf{N}_{4\times3}$ is composed of four rows of positive normal vectors on the four half-spaces. Positive normal vectors on a half-space can be acquired by cross product, $\mathbf{n}_{ij} = (K^{-1}\mathbf{l}_i) \times (K^{-1}\mathbf{l}_j)$, where $\mathbf{l}_i$, $\mathbf{l}_j$ are two adjacent enlarged corners in clockwise order to ensure the normal vector direction is inwards. If there is no existing 3D point satisfying the condition, a new set of 3D points will be registered; otherwise, the set of registered 3D points that consist of any points satisfying the condition will be updated using the following subsection method. 


\subsection{3D Point update}
Our 3D point update is adapted from the depth filter in SVO \cite{Forster2014ICRA} and developed for semantic bounding box localization. We project existing 3D points to new image frames and resample them based on the positions between their projections and the bounding boxes in the new frames (Algorithm 3). The distribution $f(\mathbf{p}|\mathbf{x})$ assesses the weights of points $\mathbf{p}$ on an image given a bounding box $\mathbf{x}$, and is formed by a combination of a Gaussian distribution and a uniform distribution:
\begin{equation}
    f(\mathbf{p}|\mathbf{x}) = w_1 h(\mathbf{p}|\mathbf{x}) + w_2 g(\mathbf{p}|\mathbf{x})
\end{equation}
where $h(\mathbf{p}|\mathbf{x})$ is distribution density function of a 2D Gaussian distribution $\mathcal{N}(\mathbf{c_x}, \Sigma_{\mathbf{x}})$, and $g(\mathbf{p}|\mathbf{x})$ is distribution density function of a 2D uniform distribution $\mathcal{U}$ among an enlarged area of the bounding box $\mathbf{x}$. Variables $w_1$ and $w_2$ are the weights of the two distributions. For $\mathcal{N}(\mathbf{c_x}, \Sigma_{\mathbf{x}})$, \\
$\mathbf{c_x}=(\frac{u_{min} + u_{max}}{2}, \frac{v_{min} + v_{max}}{2})$, $\Sigma_{\mathbf{x}} = \begin{bmatrix}
(\frac{u_{max} - u_{min}}{2})^2 & 0 \\
0 & (\frac{v_{max} - v_{min}}{2})^2
\end{bmatrix}$ \\
The 3D points $P_w$ in Algorithm 3 will be resampled according to their weights from the joint distribution by importance sampling. 

{\small 
\begin{table} 
\vspace{6pt}
\resizebox{\columnwidth}{!}{%
\begin{tabular}{l}
\hline
\textbf{Algorithm 3} 3D Point Update \hspace{32mm} \\ 
\hline
\begin{tabular}[c]{@{}l@{}}
\textbf{input}: $P_w$, $\mathbf{x}$, $T_c^w$\\ 
\textbf{output}: $P_w'$ \\
\textbf{parameter}: $\sigma$, $m$, $K$
\end{tabular}\\ 
\hline
1. \hspace{5mm} $\mathbf{a} = \sqrt{\sigma}\mathbb{N}(m, 3)$ \\
2. \hspace{5mm} $\Bar{\mathbf{a}} = S\mathbf{a} + \mathbf{b}$ \\
3. \hspace{5mm} $\mathbf{p} = [K|\mathbf{0}]T_c^w (P_w + \Bar{\mathbf{a}})$ \\
4. \hspace{5mm} $\mathbf{i} = f(\mathbf{p}|\mathbf{x})$ \\
5. \hspace{5mm} $P_w' = Importance\_Sampling(P_w, \mathbf{i})$  \\
\hline
Comments: \\
1). $\mathbb{N}(x, y)$ is an isotropic standard multi-variate Gaussian \\sampler returning a $x\times y$ matrix. \\
2). $\sqrt{\sigma}$ is the standard variance of update noise. \\
3). $f(\mathbf{p}|\mathbf{x})$ is a two-variable distribution composed of a \\Gaussian distribution and a uniform distribution. \\
\hline
\end{tabular}}
\end{table}
}
 
\subsection{Properties}
Some properties are used to describe 3D points and determine if target localization has been achieved. We apply Principle Component Analysis (PCA) to each 3D points set. The largest eigenvalue $\lambda_m$ can indicate how compact a 3D points set is. We also utilize the eigenvalues and eigenvectors from PCA to approximate an ellipsoid for the object, which can be used for visualization, to convey the information of target size, shape, and orientation. 

When a 3D points set has been updated, we want to measure the amount of information gain from the update. One metric that captures relative entropy is Kullback–Leibler divergence (KL divergence) of updated 3D points distribution with respect to previous 3D points distribution. To compute KL divergence, we approximate the 3D points set with a three-dimensional Gaussian distribution, whose mean and covariance are approximated by the mean and covariance of the 3D points. We can compute the KL divergence following the formula for multivariate Gaussian distribution:
\begin{equation}
\begin{aligned}
    D_{KL}(\mathcal{N}_0||\mathcal{N}_1) = \frac{1}{2}(tr(\Pi_1^{-1}\Pi_0) + \\(\tau_1 - \tau_0)^T\Pi_1^{-1}(\tau_1 - \tau_0) - k + \ln(\frac{|\Pi_1|}{|\Pi_0|})) 
\end{aligned}
\end{equation}
where $k=3$ is the dimension of the random variable, $\mathcal{N}_0(\tau_0, \Pi_0)$ is the estimated Gaussian distribution of the updated 3D points, $\mathcal{N}_1(\tau_1, \Pi_1)$ is the estimated Gaussian distribution of the previous 3D points, and $tr(\cdot)$ is a matrix trace function. 

Similar to bounding box tracking, we also inspect differential entropy of each 3D point set $h_{p}$. With multivariate Gaussian distribution approximation as stated above, the differential entropy shows the compactness of a 3D point set. 

Because we have no prior knowledge of targets and the UAV state estimator is not accurate, it is challenging to find a sufficient condition for localization. In practice, we jointly consider the necessary conditions such as the largest eigenvalue $\lambda_m$, KL divergence $D_{KL}(\mathcal{N}_0||\mathcal{N}_1)$, and differential entropy $h_{p}$ to decide the status of target localization. For example, $\lambda_m$ should be smaller than a threshold. When 3D points are converged, any new update will not change the 3D point distribution, and the 3D points should be compact. Thus we also need thresholds for $D_{KL}(\mathcal{N}_0||\mathcal{N}_1)$ and $h_{p}$. 

\begin{figure}[htpb]
\centering
\includegraphics[width=0.5\textwidth]{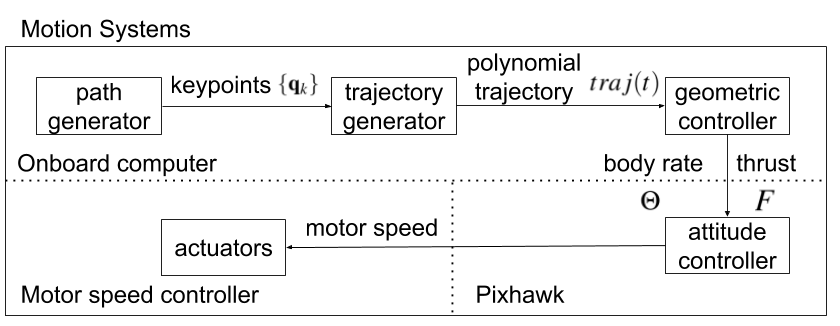}
\caption{UAV motion system.\protect\footnotemark}
\label{fig:motion_system}
\end{figure}
\footnotetext{\url{https://github.com/ZhiangChen/uav_motion}}

\subsection{Motion Planning}

We need multiple, different camera perspectives with bounding boxes to localize a target. View planning is necessary especially when there are noises in UAV state $T_c^w$ and bounding box $\mathbf{x}$. A lawn-mower flight pattern is initialized to search targets of interest. Once a set of 3D points has a largest eigenvalue $\lambda_m$ and a differential entropy $h_{p}$ that are smaller than certain thresholds but not small enough to accurately localize the target due to the lack of enough camera perspectives, we switch to a motion planning for fine localization. 

\begin{figure}
\centering
\vspace{6pt}
\includegraphics[width=0.5\textwidth]{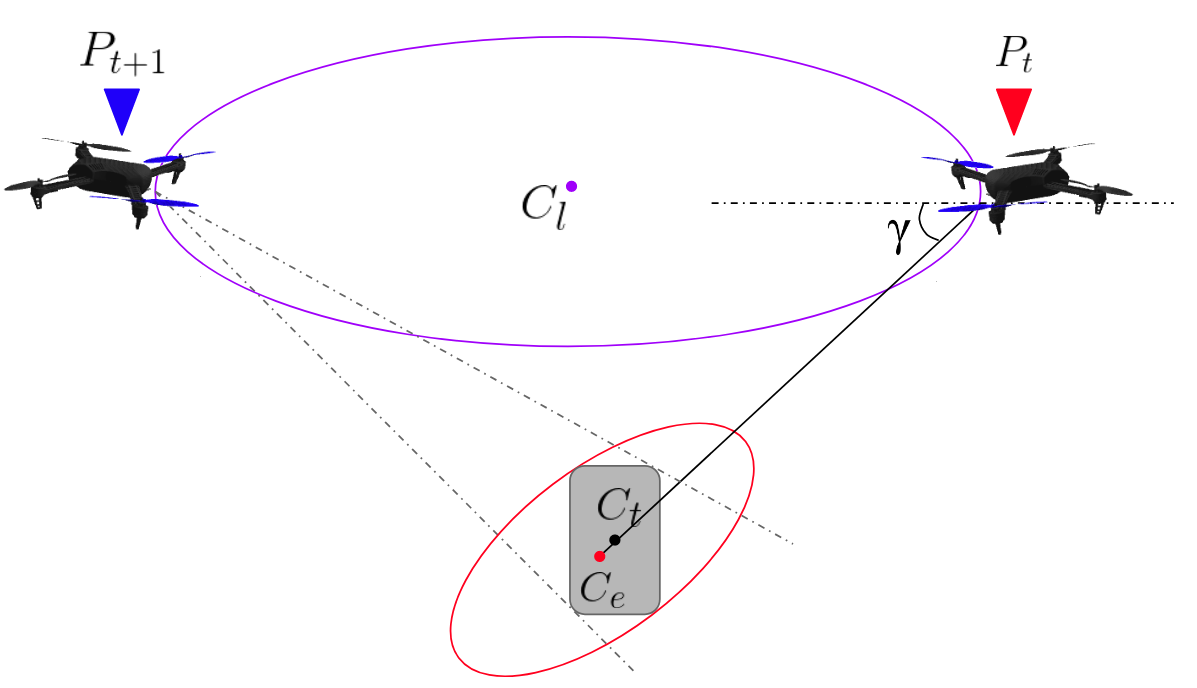}
\caption{Path planning for fine localization: The ellipse at the bottom represents a sideview of the ellipsoid from PCA. $C_e=(x_e, y_e, z_e)$ is the center of the ellipsoid. The dark-gray rounded rectangular represents a target with the center $C_t$. The large ellipse on the top is the circular path with the center $C_l=(x_e, y_e, h_s)$. $\gamma$ is the angle between the camera $z$ axis and the horizontal plane. $P_t$ is the path planning start pose; $P_{t+1}$ is an illustrated next-best-view pose.}
\label{fig:localization}
\end{figure}

We approach the motion planning problem by first considering geometric constraints and then finding a feasible solution. We want to have only one motion system in all target search, fine localization, and mapping. The motion system is illustrated in Fig.~\ref{fig:motion_system}. The trajectory generator takes keypoints from the path generator and minimizes the snap of 10th-degree polynomials \cite{richter2016polynomial}. The geometric controller \cite{jaeyoung_lim_2019_2619313, lee2010geometric} takes a desired robot's position and yaw heading and outputs desired robot's body rate and thrusts. In the target search stage, the path generator is a hard-coded lawn-mower pattern. For fine localization, the path generator should produce a set of keypoints, i.e. a set of robot poses $\{\mathbf{q}_k\}$ such that the sequential controllers can generate camera perspectives to reduce uncertainty in the target location. Because the geometric controller is only concerned with desired position and yaw, we add maximum acceleration and maximum velocity constraints in trajectory generator so that UAV's pose can be approximated to hovering. That is, UAV's roll and pitch angles will be negligible to affect the target projections in images. For view planning, we simply consider hovering pose without other UAV dynamics. The sketch of the motion planning is presented in Fig.~\ref{fig:localization}. Because the distance $d_1$ between $C_e$ and the camera is much larger than the distance $d_2$ between the camera to the UAV ($d_1:d_2>10:1$), we do not distinguish camera position or UAV position in the view planning.

We constrain the UAV positions on the horizontal plane that is parallel to $XY$ plane in world frame and has the same height as the lawn-mower search height $h_s$. There are four variables to be determined $(x, y, yaw, \gamma)$, where $(x, y)$ is the UAV coordinates on the horizontal plane in world frame, $yaw$ is the UAV's heading, and $\gamma$ is the angle between the camera $z$ axis and the horizontal plane (Fig.~\ref{fig:localization}). Because we use the 3D points information from rough localization to decide a view planning and the rough localization is not accurate, the 3D points projection in the image can deviate from the actual target projection. To alleviate this issue, we set the camera $z$ axis to pass through the 3D point center (mean Euclidean distance) $C_e$ to keep the target projection in the image frame as much as possible. Thus the UAV's heading $yaw$ is correspondingly determined. 

We decide the next-best-view pose to align the camera $z$ axis to the eigenvector with the smallest eigenvalue (smallest eigenvector) from PCA of 3D points. That is, the image plane from the next-best-view pose is parallel to the plane formed by the eigenvectors with the largest and the second largest values, which are the directions with the two largest uncertainty. From such a camera pose, the bounding box on the image plane can reduce most uncertainty in the 3D points.

When the smallest eigenvector is parallel to the horizontal plane, the next best view will be at infinity and $\gamma\to0$. We relax the optimality by fixing the angle $\gamma$ between the camera $z$ axis and the horizontal plane and minimizing the angle between camera $z$ axis and the smallest eigenvector. By having a fixed angle $\gamma$, the remaining two variables $(x, y)$ will be constrained by a circle $\mathbf{C}_l: (x-x_e)^2 + (y-y_e)^2 = \frac{(h_s - z_e)^2}{\tan^2{(\gamma)}}$ (notations are defined in Fig.~\ref{fig:localization}). Due to the ambiguity of the eigenvector direction, we set the smallest eigenvector direction to have a non-negative value along $z$ world axis $\upsilon_z$. When $\upsilon_z>0$, the solution to the minimization of the angle between camera $z$ axis and the smallest eigenvector is the intersection between the circle and the eigenvector projection on the circle plane $\mathbf{C}_l$. When $\upsilon_z=0$, we set the next-best-view point as the farthest point on the circle $\mathbf{C}_l$ to the current UAV position. We add a circular path to connect the current UAV pose and the next-best-view pose. The start keypoint is current UAV pose. The next keypoint is the closest point on the circle $\mathbf{C}_l$. Then the UAV will follow the circle curve $\mathbf{C}_l$ to approach the next-best-view pose and the heading $yaw$ is always toward the 3D points center $C_e$. 

\begin{figure}
\centering
\vspace{6pt}
\includegraphics[width=0.45\textwidth]{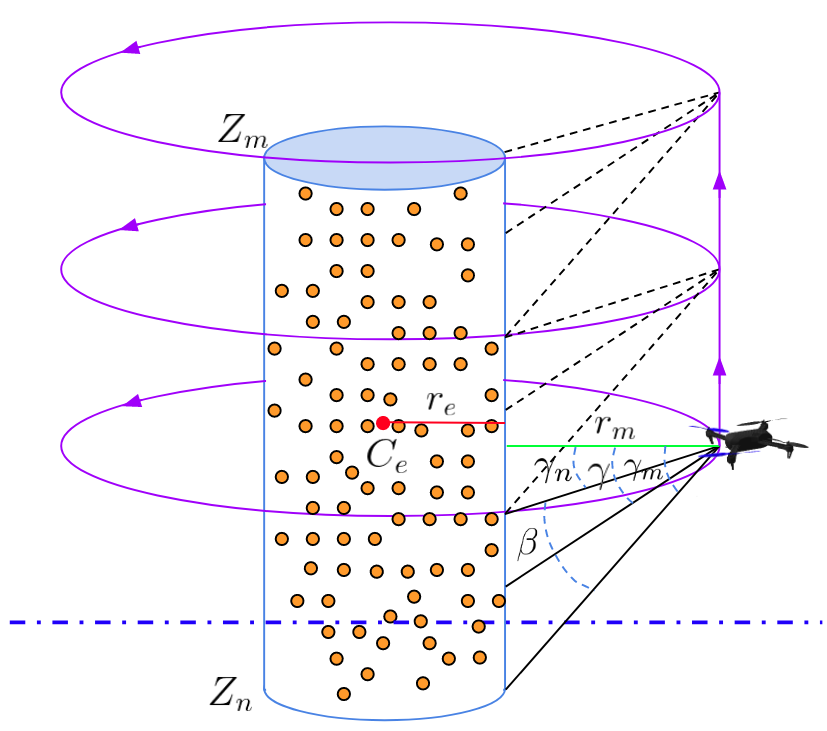}
\caption{Path planning for mapping: The small yellow dots represent 3D points of a target from fine localization. $C_e$ is the center of the 3D points (mean Euclidean distance). $r_e$ is the radius of the cylinder. $Z_m$ is the altitude of the cylinder top. $Z_n$ is the altitude of the cylinder bottom. $r_m$ is the distance to be kept between the UAV and the cylindrical curve. The vertical scanning $\beta$ is an angle slightly smaller than the angle between the upper scanning ray and the lower scanning ray of the camera. $\beta$ is similar to the concept of camera vertical field of view but is smaller. The three illustrated circles are the planned path for mapping in this case. $\gamma_m$ is the lower scanning angle $\gamma_m = \gamma - \frac{\beta}{2}$. $\gamma_n$ is the upper scanning angle $\gamma_n = \gamma + \frac{\beta}{2}$}
\label{fig:mapping}
\end{figure}

\begin{figure}
\centering
\vspace{6pt}
\includegraphics[width=0.45\textwidth]{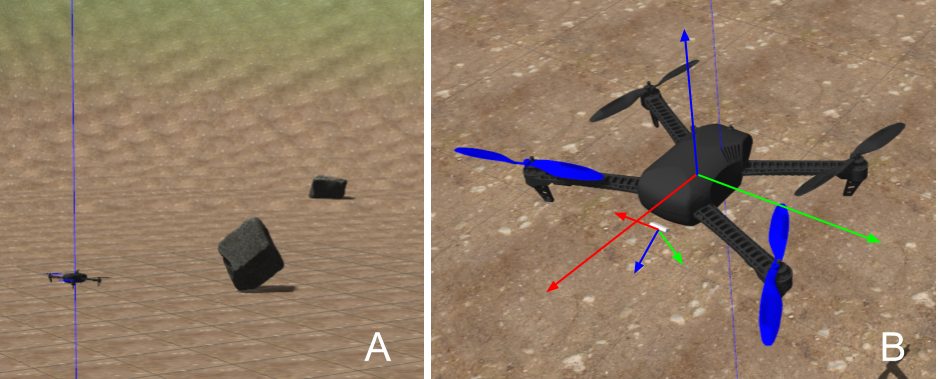}
\caption{Experiment I Gazebo simulation from DEM: (A) PBR world. (B) 3DR Iris with RGBD camera.}
\label{fig:exp1_gazebo}
\end{figure}

\begin{figure}
\centering
\includegraphics[width=0.35\textwidth]{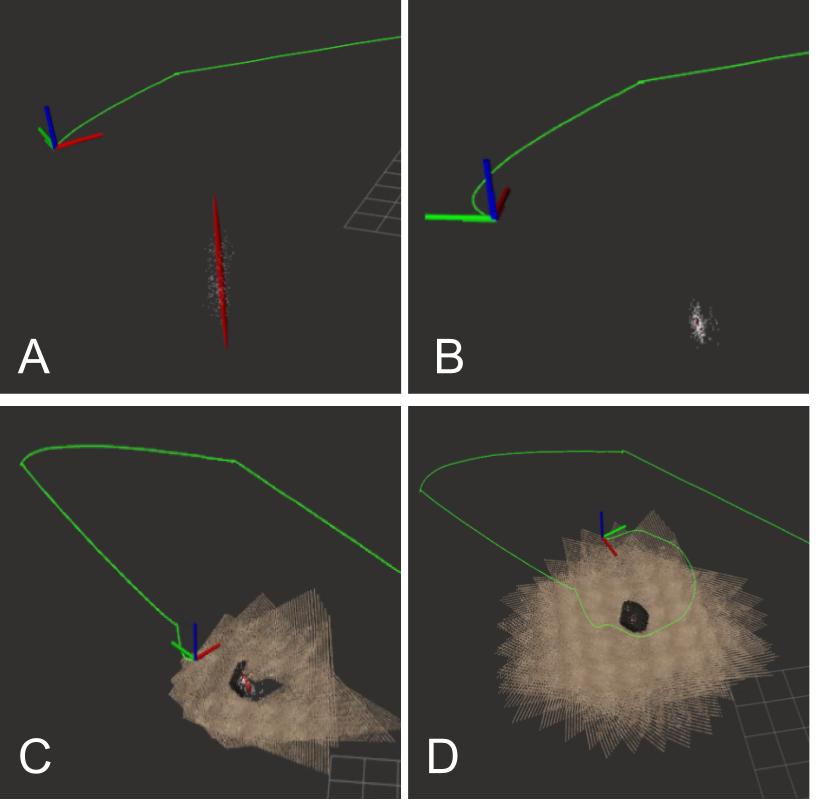}
\caption{Localization and mapping visualization in Rviz: (A, B) target localization. (C, D) target mapping. The green curves represents UAV path. The vertical red ellipsoid in (A) is formed by three eigenvectors and their corresponding eigenvalues from PCA of 3D points, which indicate uncertainties along three orthogonal axes. }
\label{fig:exp1_gazebo_result}
\end{figure}

\section{target mapping}
We deploy an RGBD SLAM \cite{labbe2019rtab} for target mapping after fine localization. We use the same motion system, except that the keypoints from the path generator should include enough camera views to densely map the full-body of a target. The 3D points from the fine localization provide occupancy information of a target, which is exploited to plan a path for the mapping. We apply a circular scanning for this purpose as shown in Fig.~\ref{fig:mapping}. We first fit the 3D points with a vertical cylinder: the cylinder's axis traverses the center of the 3D points $C_e$, its top is at the highest 3D point, its bottom is at the lowest 3D point, and its radius is the smallest radius that can include all 3D points. One objective of the path planning is to provide enough camera views to determine the occupancy certainty of the space around the cylinder.

We deploy a circular path for the mapping, which can produce 360-degree scans of the target (Fig.~\ref{fig:mapping}). The circles lie on horizontal planes that are parallel to $XY$ plane in world frame. The centers of the circles are on the axis of the cylinder. We set the circle radius $r_c =r_e+r_m$ to keep some distance between the UAV and the cylinder. The spacing and number of the circles are determined by the camera mount angle $\gamma$ and vertical scanning angle $\beta$ (Fig.~\ref{fig:mapping}) such that the target body will be fully covered. We have $\beta$ slightly smaller than the camera vertical field of view so that there are vertical overlaps between circular scans. The first circle is placed at the bottom where the cylinder bottom can be just covered by the lower scanning ray of the camera. The other circles are repeatedly arranged upward until the cylinder top can be covered by an upper scanning ray.

\begin{figure*} [t]
\centering
\vspace{6pt}
\includegraphics[width=0.9\textwidth]{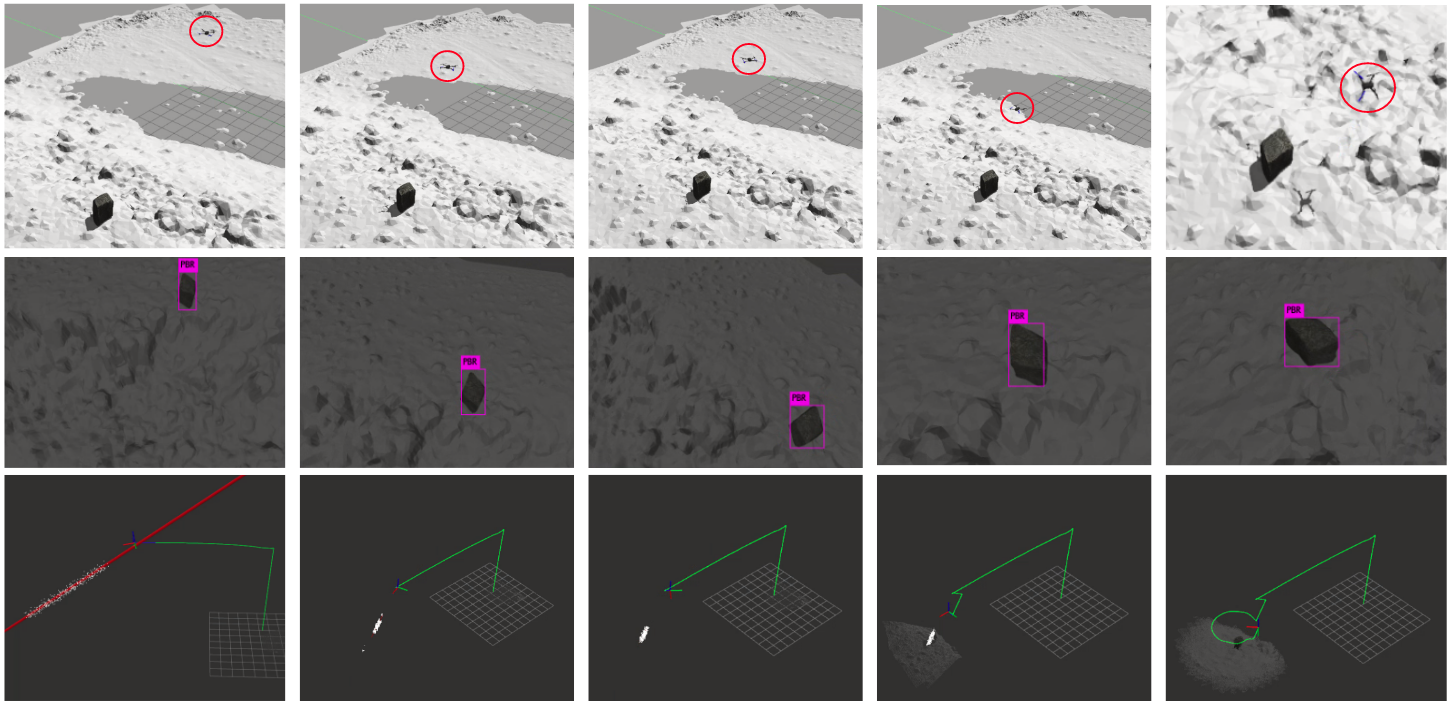}
\caption{Experiment II Gazebo simulation from Bishop fault scarp: (Top) Gazebo. (Medium) corresponding YOLO object detection. (Bottom) corresponding localization and mapping visualization in Rviz. In the top row, UAV is highlighted by red circles.}
\label{fig:exp2_bishop_result}
\end{figure*}

\begin{figure}
\centering
\vspace{6pt}
\includegraphics[width=0.45\textwidth]{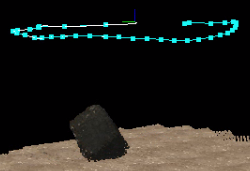}
\caption{PBR basal details from mapping: 3D map is represented by point clouds. Aqua square dots represent camera positions. }
\label{fig:exp1_pbr_base}
\end{figure}

\section{experiment}
We built the target localization and mapping system\footnotemark on ROS~\cite{ros} and conducted two experiments in Gazebo simulation using a gaming laptop (Dell G7: Intel Core i7-8750H, 16GB RAM, and Nvidia GeForce GTX 1060 Max-Q). In both two experiments, we used a 3DR Iris with an RGBD sensor ($\gamma = 55\degree$) operated by PX4 SITL and MAVROS. The UAV state estimation was maintained by an Extended Kalman Filter fusing information from IMU in the flight controller and GPS module. YOLO v2~\cite{redmon2017yolo9000} was finely tuned with the images collected in the Gazebo worlds, and the real-time inference was deployed using YOLO ROS~\cite{bjelonicYolo2018}. We applied RTAB-MAP~\cite{labbe2019rtab} for RGBD mapping, and the mapping service was only activated during the mapping stage. The vertical scanning angle $\beta$ was $40\degree$. For the motion system, the maximum velocity was $1m/s$ and the maximum acceleration was $1m/s^2$. 

\footnotetext{\url{https://github.com/ZhiangChen/target_mapping}}

\subsection{Experiment I}
We built a simple Gazebo world based on a digital elevation map (DEM) and imported 3D precariously balanced rock models into the world as shown in Fig.~\ref{fig:exp1_gazebo} (A). Lawn-mower flight patterns from various elevations were conducted, and images were collected from the camera on the 3D Iris (Fig.~\ref{fig:exp1_gazebo} (B)). We annotated 68 images in LabelMe~\cite{russell2008labelme} and finely tuned YOLO v2 that was initialized from the darknet53 model~\cite{redmon2017yolo9000}. 

The results of fine location and mapping are shown in Fig.~\ref{fig:exp1_gazebo_result}. The 3D Iris started a lawn-mower search pattern from a relatively high elevation. As a PBR was roughly localized, fine localization service was triggered and a circular motion was conducted (Fig.~\ref{fig:exp1_gazebo_result} (A)). When the 3D points were converged (Fig.~\ref{fig:exp1_gazebo_result} (B)), motion planning for target mapping was activated (Fig.~\ref{fig:exp1_gazebo_result} (C)). Then the UAV moved closer to the target and deployed RGBD SLAM (Fig.~\ref{fig:exp1_gazebo_result} (D)). 

By the presented system, we can access to the geometric information at the base of a PBR as shown in Fig.~\ref{fig:exp1_pbr_base}, which is crucial to calculate the PBR's fragility.
Such detailed basal information, however, is not available from UAS-SfM with a lawn-mower flight pattern. 

\subsection{Experiment II}
In this environment, the 3D terrain model (Fig.~\ref{fig:pbrs} (B)) was built from a fault scarp near Bishop, California, which was modeled by the previous work~\cite{chen2019geomorphological}. A 3D PBR model was placed on the foot wall of the fault scarp. We used the same vehicle and examined the presented system. As a similar process as the Experiment I, we collected aerial images from different elevations, annotated 162 images, and finely tuned YOLO v2.  

The localization and mapping processes are shown in Fig.~\ref{fig:exp2_bishop_result}. From left to right columns, the 3D Iris first tracked the PBR in image coordinates and generated 1000 3D points in the polyhedral cone. When the 3D points were roughly converged in the second column, fine localization service was requested, and the UAV started a circular motion. The fine localization service was done shortly as shown in the third column. Although the objective pose in localization stage is the next-best-view pose that aims to reduce the most uncertainty in 3D points, from both experiments, the targets could be finely localized before the UAV reached the next-best-view poses. Once the PBR was finely localized, RGBD mapping service was triggered (the fourth column), and the UAV moved close to the PBR and conducted circular motion for target mapping. The mapping service was done in the last column.


\section{contribution and future work}
We present a localization and mapping system for UAVs to map sparse geologic features such as PBRs, whose geometric fragility parameters can provide valuable information on earthquake processes and landscape development. The presented system leverages semantics and features from object detection to track targets, localizes the targets from multiple different camera perspectives using a depth filter based algorithm, and maps the targets by RGBD SLAM. Such a system provides access to autonomous PBR mapping, which was lacking in field geology but will be essential to geologic model assessment.  

The latency in target tracking increases when there are more than 8 objects tracked in the same image, which can cause false 3D point generation issue. We will alleviate this problem by using our target tracking algorithm to generate more labels offline, which will be used to retrain and improve the object detection network. The bounding box tracking will be substituted with tracking algorithms without using image features~\cite{1517Bochinski2017}. Additionally, in the fine localization stage, bundle adjustment can be used to jointly improve the accuracy of the target localization and the UAV state estimation. 

Our aim is to deploy this localization and mapping system targeting actual boulder fields with many targets and assess the quality of the final 3D target mapping, which will lead to subsequent applications to earthquake assessment and interpretation of landscape change and evolution.


\section*{ACKNOWLEDGEMENTS}
This work was supported in part by Southern California Earthquake Center (SCEC) award 20129, National Science Foundation award CNS-1521617, and National Aeronautics and Space Administration STTR award 19-1-T4.01-2855.

{\small
\printbibliography}

\end{document}